# Improving Neural Protein-Protein Interaction Extraction with Knowledge Selection

Huiwei Zhou[1], Xuefei Li[1], Weihong Yao[1,*], Zhuang Liu[1], Shixian Ning[1], ChengkunLang[1], and Lei Du[2]

[1]School of Computer Science and Technology, Dalian University of Technology, Dalian 116024, Liaoning, China

[2]School of Mathematical Sciences, Dalian University of Technology, Dalian, China 116024, Liaoning, China

* Corresponding author.

Email:zhouhuiwei@dlut.edu.cn,lixuefei@mail.dlut.edu.cn,weihongy@dlut.edu.cn,{zhuangliu1992,ningshixian,kunkun}@mail.dlut.edu.cn,dulei@dlut.edu.cn

**Abstract** Protein-protein interaction (PPI) extraction from published scientific literature provides additional support for precision medicine efforts. Meanwhile, knowledge bases (KBs) contain huge amounts of structured information of protein entities and their relations, which can be encoded in entity and relation embeddings to help PPI extraction. However, the prior knowledge of protein-protein pairs must be selectively used so that it is suitable for different contexts. This paper proposes a Knowledge Selection Model (KSM) to fuse the selected prior knowledge and context information for PPI extraction. Firstly, two Transformers encode the context sequence of a protein pair according to each protein embedding, respectively. Then, the two outputs are fed to a mutual attention to capture the important context features towards the protein pair. Next, the context features are used to distill the relation embedding by a knowledge selector. Finally, the selected relation embedding and the context features are concatenated for PPI extraction. Experiments on the BioCreative VI PPI dataset show that KSM achieves a new state-of-the-art performance (38.08% $F$1-score) by adding knowledge selection.

## 1 Introduction

The intricate networks of protein-protein interaction (PPI) contribute to controlling cellular homeostasis and the development of diseases in specific contexts. Understanding how gene mutations and variations affect the cellular interactions provides vital support for precision medicine efforts. Numerous PPIs are manually curated into structure knowledge databases (KBs) by biomedical curators, such as IntAct [1] and BioGrid [2]. However, manually extracting these PPIs from rapidly-growing biomedical literature is expensive and difficult to keep up-to-date. Automatically extracting these relations from biomedical literature is of great importance to expediting database curation.

To promote these issues, the BioCreative VI proposes a challenging task of applying text-mining methods to automatically extract protein-protein interaction affected by genetic mutations [3]. The novel challenge for the biomedical natural language processing community receives growing interests. Many automatic PPI extraction methods have been proposed [4], [5], [6], [7], [8], still, most of which focus on capturing the semantic information in the given contexts.

At the same time, KBs contain huge amounts of structured data as the form of triples (head entity, relation, tail entity) (denoted as $(h, r, t)$ ),where relation indicates the relation between the two entities. These triples could provide rich prior knowledge indicating relations between entities, which are useful for relation extraction. Nevertheless, the prior knowledge must be selectively used so that it is suitable for different contexts.

This paper proposes a Knowledge Selection Model (KSM), which consists of two modified Transformer encoders, a knowledge selector and a mutual attention, to combine the selected prior knowledge of protein-protein pairs and context information for PPI extraction. Experiments on the BioCreative VI PPI dataset show that both the selected prior knowledge and context information are effective in PPI extraction.

The major contributions of this paper are summarized as follows:

- We apply two modified Transformers to encode the context sequence of a protein pair according to each protein embedding, respectively, which could acquire entity-related dependencies of the sequence.
- A mutual attention is employed to capture the important context features towards a protein pair, and establish the connection between the two proteins.
- We devise a novel knowledge selector to distill the relation embedding based on the context features, making the prior knowledge suitable for different contexts.

## 2 Related Work

Many automatic PPI extraction methods have been proposed, which can be divided into three categories: rule-based methods, feature-based methods and neural network-based methods. Rule-based methods [4] are simple and effective, but hard to apply to a new dataset. Chen et al. [4] use a simple rule-based co-occurrence strategy, and get an $F1$-score of 27.4% on the BioCreative VI PPI dataset. Feature-based methods [4], [5], [6] extract PPIs based on one-hot represented lexical and syntactic features. Chen et al. [4] also use support vector machine with graph kernel, obtaining 33.66% $F1$-score on the BioCreative VI PPI dataset. Generally, feature-based methods rely heavily on suitable features, which require extensive feature engineering.

Recently, neural network-based methods have been proposed to learn low-dimensional semantic representations of word sequences for relation extraction without manual feature engineering. Zeng et al. [9] first employ Convolutional Neural Network (CNN) [10] to learn distributed representations for relation extraction. Their method achieves better performance than feature-based methods. As for BioCreative VI PPI extraction task, Tran and Kavuluru [7] employ CNN to extract local semantic features and get an $F1$-score 36.33%. Rios et al. [11] use CNN as a base model, and then improve the model with unlabeled data via an adversarial process. They finally improve $F1$-score to 36.77%. CNN pays more attention to local features by performing convolutions within the varying filter windows, but neglects the long-range dependencies.

Some efforts have been made to capture long-term dependencies by using recurrent neural network (RNN) [12], long short-term memory network (LSTM) [13] or memory network models [14], [15]. Wang et al. [8] use RNN to model long-range contextual dependencies, achieving an $F1$-score of 16.43% on the BioCreative VI PPI dataset. Zhou et al. [16] adopt memory network for PPI extraction, and show that an external long-term memory is superior to the local memories in LSTM. They get an $F1$-score of 32.38% with only contextual information. These models could capture long-term dependencies, but could not conduct direct connections between two arbitrary tokens in a sequence.

Recently, attention mechanisms are extensively added to RNN and CNN architectures to focus on the important contextual information [17], [18]. Attention mechanisms allow RNN and CNN to explicitly calculate how much each element in the input sequence contributes to sequence representations. However, such attention mechanisms are usually based on complex recurrence or convolutional networks.

To dispense with recurrence and convolutions, a simple network based solely on attention mechanisms, called Transformer, is proposed for neural machine translation tasks and achieves a state-of-the-art performance [19]. Transformer relies entirely on self-attention and multi-head attention to compute direct dependencies between any two context words in a sequence. Due to its superior performance, Transformer is used to pre-train language models, such as OpenAI GPT-2 [20] and BERT [21], which can be fine-tuned with just one additional output layer and obtain state-of-the-art results on a wide range of tasks.

On the other hand, large-scale KBs usually store prior knowledge in the form of triplet $(h,r,t)$. The prior knowledge in KBs is effective for relation extraction. Previous work expresses prior knowledge as one-hot features to improve relation extraction performance [22], [23]. However, such one-hot features assume that all the objects are independent of each other. Recently, knowledge representation learning methods [24], [25], [26], [27], [28] have been proposed to encode the entities and relations in KBs into low-dimensional vectors, which could find the potential semantic relations between entities and relations. Among these methods, TransE [24] is simple but can achieve the state-of-the-art predictive performance. Zhou et al. [16] leverage both entity embeddings and relation embeddings learned by TransE for PPI extraction, and improve the $F$1-score from 32.38% to 35.91% on the BioCreative VI PPI dataset. TransE-based knowledge representations are easy to combine with neural networks, like blindly concatenating as features. Nevertheless, knowledge should be distilled based on contexts.

To capture direct dependencies between any two words in a sequence and selectively introduce prior knowledge, this paper proposes a Knowledge Selection Model (KSM) based on Transformer encoder for PPI extraction.

## 3 Method

KSM uses two Transformers and a mutual attention to extract the important context features related to protein pairs, which are then fed to a knowledge selector to distill the prior knowledge. Both the important context features and the selected prior knowledge are used for PPI extraction. In this section, we first describe the preprocessing procedure and feature representation to ease the exposition, and then elaborate details of KSM.

### 3.1 Preprocessing and Feature Representation

Since PPI relations are annotated at the document level in the BioCreative VI PPI dataset [29], the context word sequences of each protein pair in a document are considered as candidate instances. To reduce the number of inappropriate instances, the sentence distance between a protein pair should be less than 3. Otherwise, the protein pair will not be considered. The context word sequence with respect to a protein pair consists of words between the protein pair and three expansion words on both sides. To simplify the interpretation, we assume the mentions of a protein pair are two single words $C_{i_1}$ and $C_{i_2}$, where $i_1$ and $i_2$ are the positions of the two protein entities. For a given text $\{...,c_{i_1-3},c_{i_1-2},c_{i_1-1},c_{i_1},c_{i_1+1}...,c_i,...,c_{i_2-1},c_{i_2},c_{i_2+1},c_{i_2+2},c_{i_2+3},...\}$, the context word sequence can be expressed as $\{c_{i_1-3},c_{i_1-2},c_{i_1-1},c_{i_1+1}...,c_i,...,c_{i_2-1},c_{i_2+1},c_{i_2+2},c_{i_2+3}\}$. As can be seen, the mentions of the focused protein pair in the current instance are removed. Besides, all the other protein mentions are replaced with "gene0". The numbers in the context are replaced by a specific string "NUMBER". Some special characters, such as "*", are removed. In the training phase, if a protein pair is annotated as an interaction pair in a document, each instance of the protein pair in the document is labeled as a positive instance, otherwise a negative instance. In the test phase, a protein pair is recognized as an interaction pair as long as one of its candidate instances is classified as positive.

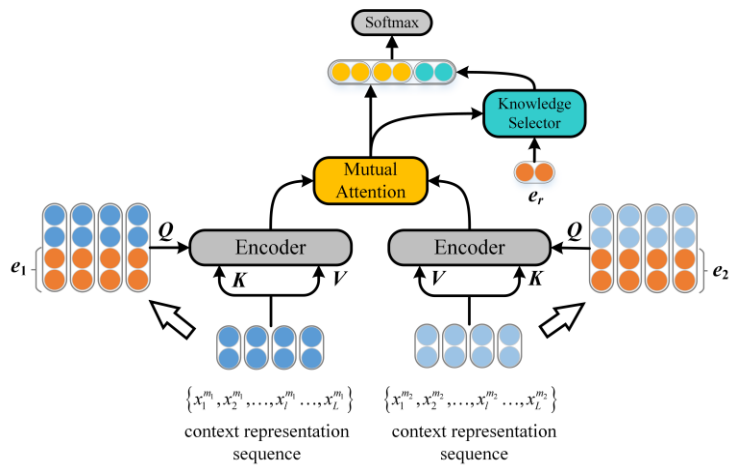

Fig.1. The structure of the knowledge selection model. It consists of three components: two modified Transformer encoders, a mutual attention and a knowledge selector. Each encoder learns entity-related long-range dependencies of the context, according to each entity embedding $e_1$ / $e_2$ of a protein pair. The mutual attention is applied on the two encoded sequences to generate the context features related to the protein pair. The knowledge selector aims at to distill the relation embedding $e_r$.

The input of KSM consists of two parts, the context word sequence and prior knowledge of the protein pair. We transform the words in the context sequence into continuous vectors, word embeddings, preserving both syntactic and semantic information well [30]. Thus, the context of the protein pair is represented as a sequence of word embeddings $\{w_1, w_2, ..., w_l, ..., w_L\}$, where $w_l \in \mathbb{R}^d$ and $L$ is the length of the context.

The entity-relation triples in KBs are considered as prior knowledge. Entities and their relations in entity-relation triples $(m_1, r, m_2)$ are embedded in a continuous vector space $\mathbb{R}^{d_{kb}}$ via TransE model, represented as $(e_1, e_r, e_2)$, where the relation embeddings $e_r$ corresponds to a translation between the embeddings of the protein pair $(e_1, e_2)$, i.e. $e_1 + e_r \approx e_2$.

Furthermore, we incorporate position information through position embeddings to reflect the relative distance from the words to the two protein mentions. Position embeddings have been shown to be effective in boosting relation classification performance [9]. The position of each context word is defined as the absolute distance from it to the protein mention. Since there are two entities $(m_1, m_2)$, two sequences of position embeddings are obtained, which are represented as $\{p_1^{m_1}, p_2^{m_1}, ..., p_l^{m_1}, ..., p_L^{m_1}\}$ and $\{p_1^{m_2}, p_2^{m_2}, ..., p_l^{m_2}, ..., p_L^{m_2}\}$, respectively. The position embeddings have a dimension size of $d$, same as word embeddings. Following Vaswani et al. [19], we encode the position to the position embedding, and add it to the corresponding word embedding which can be formalized as follows:

$$x_l^{m_i} = w_l + p_l^{m_i} \quad (1)$$

where $i \in \{1, 2\}$ and $m_i$ represents one of the two protein entities. Thus, we obtain two context representation sequences, $\{x_1^{m_1}, x_2^{m_1}, ..., x_l^{m_1}, ..., x_L^{m_1}\}$ and $\{x_1^{m_2}, x_2^{m_2}, ..., x_l^{m_2}, ..., x_L^{m_2}\}$.

3.2 Knowledge Selection Model (KSM)

This section describes the details of KSM. The overall structure of KSM is shown in Fig. 1, which has three components, two modified Transformer encoders, a mutual attention and a knowledge selector.

Firstly, two sequences $\{x_1^{m_1}, x_2^{m_1}, ..., x_l^{m_1}, ..., x_L^{m_1}\}$ and $\{x_1^{m_2}, x_2^{m_2}, ..., x_l^{m_2}, ..., x_L^{m_2}\}$ along with two entity embeddings $e_1$ and $e_2$ are the inputs of two encoders, respectively. We anticipate that the modified Transformer encoder could learn entity-related long-range dependencies of the context. Next, the outputs of the two sequence encoders are fed into the mutual attention mechanism to establish the connection between the two entities and extract important context features with regard to the protein pair. Finally, the context features are used to distill relation embeddings by the knowledge selector. The selected relation embeddings and the context features are concatenated for relation extraction. In following subsections, we describe the three components in detail.

*3.2.1 Sequence Encoder*

The sequence encoder of our model is built upon the Transformer encoder. Transformer originally has an encoder-decoder architecture and is firstly applied in machine translation task [19]. Our sequence encoder consists of *N* blocks, each of which contains two sublayers: a multi-head attention layer and a position-wise feed-forward layer.

(1) Multi-head attention builds upon scaled dot product attention for mapping a query and a set of key-value pairs to an output:

$$\text{Attention}(Q, K, V) = \text{softmax}(\frac{QK^T}{\sqrt{d_K}})V \quad (2)$$

where the query ($Q$), keys ($K$), values ($V$) and output are all list of vectors with equal length and $d_K$ is the dimension of $K$. Vaswani et al. [19] point out that the input of softmax grows large in magnitude, pushing the softmax function into regions where it has extremely small gradients. Therefore, the dot productions are scaled by $1/\sqrt{d_K}$ to counteract this effect.

Instead of performing a single attention function with $d$-dimension $Q$, $K$ and $V$, the multi-head attention performs the scaled dot product attention multiple times on linearly projected $Q$, $K$ and $V$. The multi-head self-attention allows the model to jointly attend to information from different representation subspaces at different positions [19]. In this paper, entity embeddings learned from KBs are introduced so that

Transformer encoders could take the prior knowledge of protein embeddings into consideration. Specifically, we concatenate the entity embedding to each position of input sequence, forming the Q of each block. There are two entities $e_1$ and $e_2$ for the two sequence encoders, respectively. Here we take $e_1$ as an example, and Q, K, V are as follows:

$$Q = \{[x_1^{m_1}, e_1], [x_2^{m_1}, e_1], \ldots, [x_l^{m_1}, e_1] \ldots, [x_L^{m_1}, e_1]\}$$
$$K = \{x_1^{m_1}, x_2^{m_1}, \ldots, x_l^{m_1} \ldots, x_L^{m_1}\} \quad (3)$$
$$V = \{x_1^{m_1}, x_2^{m_1}, \ldots, x_l^{m_1} \ldots, x_L^{m_1}\}$$

where $Q \in \mathbb{R}^{L \times (d+d_{kb})}$, $K \in \mathbb{R}^{L \times d}$ and $V \in \mathbb{R}^{L \times d}$. Thus, the multi-head attention with prior knowledge can be formulated as:

$$V^{MH} = [head_1, head_2, \cdots, head_h] W^H$$
$$head_i = \text{Attention}(QW_i^Q, KW_i^K, VW_i^V) \quad (4)$$

where $W_i^Q \in \mathbb{R}^{(d+d_{kb}) \times d_{head}}$, $W_i^K \in \mathbb{R}^{d \times d_{head}}$, $W_i^V \in \mathbb{R}^{d \times d_{head}}$ and $W^H \in \mathbb{R}^{hd_{head} \times d}$ are trainable projection matrices, and $[head_1, head_2, \cdots, head_h]$ is a concatenation of outputs of $h$ heads.

In this way, we expect to effectively compute dependencies between any two context words and encode the prior knowledge of the protein embedding into each sequence word.

(2) The second sublayer in each encoder block is a position-wise feed-forward layer defined as:

$$V^{FF} = \text{ReLU}(V^{MH} W_1 + b_1) W_2 + b_2 \quad (5)$$

where $W_1 \in \mathbb{R}^{d \times d}$, $W_2 \in \mathbb{R}^{d \times d}$, $b_1 \in \mathbb{R}^d$ and $b_2 \in \mathbb{R}^d$ are trainable parameters. The output $V^{FF}$ forms K and V of next block, while we concatenate entity embedding $e_1$ to each position of $V^{FF}$ to form Q of next block.

For each sublayer, a residual connection [31] followed by layer normalization [32] is applied. For regularization, dropout [33] is inserted before residual connections.

Finally, $V^{FF}$ of the last position-wise feed-forward layer forms the encoded sequence $V^{m_1} \in \mathbb{R}^{L \times d}$, as the output of a sequence encoder. As for $e_2$, we also get the corresponding $V^{m_2} \in \mathbb{R}^{L \times d}$.

### 3.2.2 Mutual Attention

To establish the connection between the two entities, we adopt a mutual attention on the two encoded sequences $V^{m_1}$ and $V^{m_2}$ to generate the important context features related to the protein pair. Our mutual attention is based on the additive attention or multi-layer perceptron attention, which is firstly introduced in machine translation task [34] and conversation generation task [35]. Specifically, for each pair of $V_i^{m_1}$ and $V_j^{m_2}$ (i.e. the $i$-th row of $V^{m_1}$ and the $j$-th row of $V^{m_2}$) in the two sequences, we compute the attention score as follows:

$$\alpha_{ij} = w^T \tanh(W^1 V_i^{m_1} + W^2 V_j^{m_2}) \quad (6)$$

where the $W^1 \in \mathbb{R}^{d \times d}$, $W^2 \in \mathbb{R}^{d \times d}$ and $w \in \mathbb{R}^{d \times 1}$ are the weight matrices. We then compute the attention vector $s^{m_1}$ for entity $m_1$:

$$\beta_i^{m_1} = \frac{1}{L} \sum_{j=1}^{L} \alpha_{ij}$$
$$p_i^{m_1} = \frac{\exp(\beta_i^{m_1})}{\sum_{k=1}^{L} \beta_k^{m_1}} \quad (7)$$
$$s^{m_1} = \sum_{i=1}^{L} p_i^{m_1} V_i^{m_1}$$

The average score in each row of $\alpha$ are fed to a softmax function, achieving normalized weight $p_i^{m_1}$ for each $V_i^{m_1}$. Then, $s^{m_1} \in \mathbb{R}^d$ is computed as weighted sum of $V_i^{m_1}$.

Similar to $s^{m_1}$, we compute the average score $\beta_j^{m_2}$ in each row of $\alpha$, and get the weighted sum vector $s^{m_2} = \sum_{j=1}^{L} p_j^{m_2} V_j^{m_2}$ for entity $m_2$, where $p_j^{m_2}$ is the normalized weight for each $V_j^{m_2}$.

Eventually, we concatenate $s^{m_1}$ and $s^{m_2}$ as the context features, which are used in the following knowledge selector and relation classification.

### 3.2.3 Knowledge selector

The relation embeddings learned from KBs are deemed as crucial clue for PPI extraction. We can directly concatenate the relation embeddings and context features as the final features to enhance the relation classification performance. However, this way of introducing relation embeddings might be blind. For example, there is a sentence "*…several debilitating diseases are linked to heritable mutations in **RyR1** and **RyR2** including malignant hypothermia, central core disease…*" in PPI dataset. Apparently, the sentence does not describe interaction between the entity pair "**RyR1**" and "**RyR2**", although we can extract the interaction relation of this pair from KBs. In such case, blindly concatenating the relation embeddings as features could cause conflicts between prior knowledge and contexts.

To alleviate this problem, we want a knowledge selector to distill the prior knowledge according to what context information and prior knowledge are in hand. Thus, both context features and relation embeddings should be considered. The knowledge selector is expected to achieve a weight vector that has equal length with relation embeddings, so that relation embeddings could be distilled through an element-wise operation (Hadamard product is used in KSM).

With this in mind, we adopt a gate-based method as a knowledge selector, whose inputs are relation embeddings and context features, and the output are the selected relation embeddings. The details of the knowledge selector are as follows:

$$g = activation(W^{ks}[s^{m_1}, s^{m_2}] + U^{ks}e_r + b^{ks})$$
$$e_r^{ks} = elementWiseOp(g, e_r) \quad (8)$$

where $W^{ks} \in \mathbb{R}^{d \times 2d}$, $U^{ks} \in \mathbb{R}^{d \times d}$, $b^{ks} \in \mathbb{R}^d$ are trainable parameters of knowledge selector. *activation* is a non-linear activation function like sigmoid, ReLU, or hyperbolic tangent. *elementWiseOp* is an element-wise operation. We explore the effects of different activation function in the experiments.

The selected relation embeddings and context features are concatenated as the final features $[s^{m_1}, s^{m_2}, e_r^{ks}] \in \mathbb{R}^{3d}$ to perform relation classification.

*3.2.4 Classification and Training*

We feed $[s^{m_1}, s^{m_2}, e_r^{ks}]$ into a fully-connected layer followed by a softmax layer for relation classification.

$$p(y = j | I) = \text{softmax}(W_s[s^{m_1}, s^{m_2}, e_r^{ks}] + b_s)$$
$$\hat{y} = \underset{y \in [0,1]}{\arg\max}(p(y = j | I)) \quad (9)$$

where $\hat{y}$ is our prediction, $W_s \in \mathbb{R}^{2 \times 3d}$ and $b_s \in \mathbb{R}^2$ are trainable parameters, and $I$ is the training instances. The loss function is defined as follows:

$$J(\theta) = -\frac{1}{N}\sum_{i=1}^{N} \log p(y_i | I_i, \theta) \quad (10)$$

where $N$ is the number of labelled instances in the training set, $I_i$ is the $i$-th instance, $y_i$ is the golden label, and $\theta$ is the parameters of the entire model.

## 4 Experiments and Results

4.1 Dataset and Evaluation Metrics

**Dataset.** Experiments are conducted on the BioCreative VI Track 4 PPI extraction task dataset ("PPI dataset" for short) [29]. The organizers provide 597 PubMed abstracts for training and 1,500 abstracts for test. Protein entities in the training and test sets are recognized by GNormPlus [36] toolkits (https://www.ncbi.nlm.nih.gov/CBBresearch/Lu/Demo/tmTools/download/GNormPlus/GNormPlusJava.zip), and normalized to Entrez Gene ID. According to Chen et al. [4], GNormPlus have a recall of 53.4% on the training set, which means not all protein mentions are annotated. For training set, we can simply put the missed protein mentions back based on the golden annotations provided by the training set, and then generate the training protein pairs. But for test set, we have no evidence to get the missed protein mentions back. Table 1 lists the statistics of the training set, test set and test set annotated by GNormPlus ("Test-G" for short).

**Table 1.** Statistics of the BioCreative VI PPI dataset.

| Dataset | #Article | #PPI |
|---|---|---|
| **Train** | 597 | 752 |
| **Test** | 1500 | 869 |
| **Test-G** | 1500 | 483 |

"#Abstract" and "#PPI" mean the number of abstracts and protein-protein interaction affected by mutations in datasets, respectively.

We extract PPI relation triples from two knowledge bases, IntAct (https://www.ebi.ac.uk/intact/) and BioGrid (https://thebiogrid.org/), which have the same 45 kinds of relation types. Finally, 1,518,592 triples and 84,819 protein entities are obtained for knowledge representation learning. All the protein entities in KBs are linked to the Entrez Gene IDs by using UniProt (https://www.uniprot.org/) [37] database.

**Evaluation Metrics.** For PPI task, organizers employ two level evaluation:

Exact Match: All system predicted relations are checked against the manual annotated ones for correctness.

HomoloGene Match: All gene identifiers in the predicted relations and manually annotated data are mapped to common identifiers representing HomoloGene classes, then all predicted relations are checked for correctness.

We use Exact Match evaluation to measure the overall performance of our method. Since HomoloGene Match accounts for homologous genes, that is, only when the predicted Gene ID and the annotated Gene ID are homologous genes, they are counted as a match. And most of the existing works evaluate on Exact Match rather than HomoloGene Match. The evaluation of PPI extraction is reported by official evaluation toolkit (https://github.com/ncbi-nlp/BC6PM), which adopts micro-averaged [38] Precision ($P$), Recall ($R$) and $F$1-score ($F$) based on Exact Match.

4.2 Experimental Setup

Word2Vec tool (https://code.google.com/p/word2vec/) [30] is used to pre-train word embeddings on the datasets (about 9,308MB, 27 million documents, 3.4 billion tokens and 4.2 million distinct words) downloaded from PubMed (http://www.ncbi.nlm.nih.gov/pubmed/). The dimensions of word, entity, position and relation embeddings are all $d$=100. The sequence encoder has two blocks, each of which employs $h$=4 parallel attention layers, or heads. For each head, the dimension of the projected $Q$, $K$ and $V$ is $d_{head}=d/h=25$. Thus, the dimension of input and output of FFN is 100. The model is trained by using Adadelta technique [39] with a learning rate 0.02 and a batch size 64. The whole framework is developed by PyTorch (http://pytorch.org/).

For knowledge representation learning, we initialize the entity embeddings with the averaged embeddings of words contained in entity mention and the relation embeddings with a normal distribution. The link of the TransE code is listed at footnote (https://github.com/thunlp/Fast-TransX).

4.3 Baselines

In the experiment, we compare the proposed **KSM** (Hyperbolic tangent and Hadamard product are used as *activation* and *elementWiseOp*, respectively) with the following baseline methods:

**CNN**: This method applies a classic CNN. The context sequences fed to **CNN** are the same as **KSM**. The representation of each position in a context sequence is the concatenation of a word embedding, two position embeddings (initialized randomly with dimension 10) and two entity embeddings, which can be formulated as follows:

$$x_l = [w_l, p_l^{m_1}, p_l^{m_2}, e_1, e_2] \quad (11)$$

In the convolution layer, 100 filters with different window sizes $k=\{3,4,5\}$ respectively are used to produce 300 feature maps. Then, a max-pooling is used to generate context features. Finally, the context features are concatenated with relation embeddings for relation classification.

**CNN+KS**: The structure of this model is the same as **CNN**, except that a knowledge selector (KS for short) is added to distill relation embeddings.

**BiLSTM**: This method applies a bidirectional LSTM. The context sequences fed to **BiLSTM** is the same as **KSM**. For the forward LSTM, the representation of each position in a context sequence is the concatenation of a word embedding, a position embedding (initialized randomly with dimension 10) and an entity embedding, which can be formulated as follows:

$$\vec{x}_l = [w_l, p_l^{m_1}, e_1] \quad (12)$$

For the backward LSTM, the representation is similar:

$$\overleftarrow{x}_l = [w_l, p_l^{m_2}, e_2] \quad (13)$$

The forward and backward LSTMs each have 100 hidden units. Hidden states at each time step $l$ from two directional LSTMs, $\vec{h}_l$ and $\overleftarrow{h}_l$, are concatenated and fed to an attention layer:

$$\alpha_l^{lstm} = v^T \tanh(W^{lstm}[\vec{h}_l, \overleftarrow{h}_l] + b^{lstm})$$
$$p_l^{lstm} = \frac{\exp(\alpha_l^{lstm})}{\sum_{k=1}^{L} \alpha_k^{lstm}} \qquad (14)$$
$$s_l^{lstm} = \sum_{l=1}^{L} \alpha_l^{lstm}[\vec{h}_l, \overleftarrow{h}_l]$$

Then, $s^{lstm}$ is concatenated with relation embeddings for classification.

**BiLSTM+KS**: The structure of this model is the same as **BiLSTM**, except that a knowledge selector is added to distill relation embeddings.

**KSM-KS**: The structure of this model is the same as **KSM**, except that relation embeddings are directly concatenated with the context features without knowledge selection.

**BERT**: This method applies the well-known pre-trained language model BERT [21] to encode context sequences. Similar to **KSM**, two BERT encoders are employed for a protein pair. The input sequences of two encoders are formulated as $\{x_{[CLS]}, x_1, ..., x_l, ..., x_L, x_{[SEP]}\}$. The representation of each position $x_l$ can be formulated as $x_l^{m_i} = \hat{w}_l + \hat{p}_l^{m_i}$ where $i \in \{1,2\}$, $m_i$ represents one of the two protein entities, $\hat{w}_l$ and $\hat{p}_l$ are word embeddings and position embeddings in BERT.

The two final layer outputs in the position [CLS] are concatenated with entity embeddings and relation embeddings for classification.

In our experiments, we use BERT-base-uncased with 12-layer, 768-hidden, 12-heads and 110M parameters, which is pre-trained on English news text.

**BERT+KS**: The structure of this model is the same as **BERT**, except that a knowledge selector (KS for short) is added to distill relation embeddings.

**GPT-2**: This method applies the well-known pre-trained language model GPT-2 [22] to encode context sequences. Similar to **KSM**, two GPT-2 encoders are employed for a protein pair. The context sequences fed to **GPT-2** is the same as **KSM**. The representation of each position in a context sequence is the sum of a word embedding and a position embedding, which be formulated as $x_l^{m_i} = \tilde{w}_l + \tilde{p}_l^{m_i}$ where $i \in \{1,2\}$, $m_i$ represents one of the two protein entities, $\tilde{w}_l$ and $\tilde{p}_l$ are word embeddings and position embeddings in GPT-2.

Then, a max-pooling layer is added on the top of two GPT-2 encoders to get the two results, which are concatenated with entity embeddings and relation embeddings for classification.

In our experiments, we use OpenAI GPT-2 Small-sized English model with 12-layer, 768-hidden, 12-layer, 768-hidden, 12-heads and 117M parameters, which is pre-trained on English news text.

**GPT-2+KS**: The structure of this model is the same as **GPT-2**, except that a knowledge selector (KS for short) is added to distill the relation embeddings.

**Table 2.** Comparison with respect to baselines and knowledge selection on PPI dataset.

| Methods | P (%) | R (%) | F (%) |
|---|---|---|---|
| CNN | 29.46 | 40.51 | 34.11 |
| CNN+KS | 31.16 | 40.16 | 36.15 |
| BiLSTM | 31.98 | 38.20 | 34.82 |
| BiLSTM+KS | 32.45 | 37.97 | 34.99 |
| BERT | 30.51 | 31.07 | 30.79 |
| BERT+KS | 31.13 | 30.96 | 31.04 |
| GPT-2 | 27.26 | 34.98 | 30.65 |
| GPT-2+KS | 31.32 | 30.96 | 31.13 |
| KSM-KS | 35.70 | 38.78 | 37.18 |
| KSM | 38.78 | 37.40 | **38.08** |

**+KS** indicates the knowledge selector is added to the method. **-KS** indicates the knowledge selector is removed from the method. The highest $F$1-score is marked in bold, similarly hereinafter.

The results of baselines are shown in Table 2. From the results, we can conclude the follows:

(1) **CNN+KS**, **BiLSTM+KS**, **BERT+KS**, **GPT-2+KS** and **KSM** all benefit from the knowledge selector. Especially for **CNN+KS**, 2.04% increase in $F$1-score is achieved. **CNN+KS** beats **BiLSTM+KS**, which illustrates that knowledge selector might be more suitable for non-recursive model. **KSM-KS** succeeds in capturing the long-rang entity-related dependency information and gets a higher $F$1-score than CNN-based models and LSTM-based models. With the help of knowledge selector, **KSM** provides the highest $F$1-score.

(2) Although **BERT** and **GPT-2** obtain state-of-the-art re-

sults on many NLP tasks of common domains, they perform poorly in PPI extraction. This shows that the pre-trained language models do not work well in the biomedical domain with simple fine-tuning.

4.4 Influence Factors of Knowledge Selector

To verify the effectiveness of each part of the knowledge selector, we explore three factors that could affect the performance, (1) whether the relation embeddings are involved to form the gate $g$, (2) which *elementWiseOp* is used and (3) which *activation* is used.

For the first factor, we develop **KSM-R**. **KSM-R** has the same structure as **KSM**, except that the gating unit in the knowledge selector only uses context features without relation embeddings. Thus, in **KSM-R,** the knowledge selector is modified as follows:

$$g = activation(W^{ks}[s^{m_1}, s^{m_2}] + b^{ks})$$
$$e_r^{ks} = g \odot e_r \qquad (15)$$

For the second factor, we develop **KSM(sum)**. **KSM(sum)** has the same structure as **KSM** except that the *elementWiseOp* of knowledge selector is the element-wise summation operation:

$$e_r^{ks} = g + e_r \qquad (16)$$

For the third factor, we make **KSM-R**, **KSM** and **KSM(sum)** each apply three kinds of *activation*, namely ReLU, sigmoid and tanh, to explore the effects of *activation* in the knowledge selector. The $F1$-scores of all combinations are shown in Table 3. From the results, we can conclude the follows:

(1) In general, with the relation involved, **KSM** and **KSM(sum)** outperform **KSM-R**. This demonstrates that relation embeddings are important for the knowledge selector.

(2) **KSM(sum)** uses element-wise summation as its *elementWiseOp*. This kind of knowledge selector has a similar fashion to residual networks [31], whose input $e_r$ is added to the output $g$. From Table 3, although **KSM(sum)** beats **KSM** with ReLU *activation*, **KSM** succeeds when using tanh and sigmoid *activation*.

(3) **KSM-R** and **KSM** both use Hadamard product as their *elementWiseOp*. In this case, $g$ is a gating unit, controlling the flow of relation embeddings. ReLU does not have upper bound on positive inputs but strictly zero on negative inputs, which acts just as breaking rather than gating relation embeddings. Sigmoid and tanh both scale their inputs to a fixed range, which is more compact. From Table 3, we can conclude that sigmoid and tanh are more suitable to control the magnitude of relation embeddings than ReLU.

In summary, the knowledge selector need relation embeddings and tanh *activation* to calculate $g$. Based on this, **KSM** then uses a Hadamard product as *elementWiseOp*, which provides the highest $F1$-score.

**Table 3.** Effects of each part of knowledge selector, including relation embeddings (**KSM-R** vs. **KSM**), *elementWiseOp* (**KSM** vs. **KSM(sum)**) and *elementWiseOp* (ReLU, sigmoid and tanh).

| Methods | *elementWiseOp* | | |
| --- | --- | --- | --- |
| | ReLU | sigmoid | tanh |
| **KSM-R** | 34.73 | 36.74 | 37.20 |
| **KSM** | 38.78 | 37.40 | **38.08** |
| **KSM(sum)** | 37.60 | 36.14 | 37.23 |

-R indicates the gating unit in the knowledge selector only uses context features without relation embeddings. We only show the $F1$-score (%) of all the variants.

4.5 Entity Knowledge Selection vs. Relation Knowledge Selection

A knowledge selector can also be used for the entity embedding selection. Therefore, we put knowledge selector on relation embeddings, entity embeddings and both to compare the performances.

For the knowledge selector of entity embeddings, we average $V$ as an input. Take $m_1$ as an example:

$$g = activation\left(W^{kse}\left(\frac{1}{L}\sum_{i=1}^{L} x_i^{m_1}\right) + U^{kse}e_1 + b^{kse}\right)$$
$$e_1^{kse} = elementWiseOp(g, e_1) \qquad (17)$$

where the $W^{kse} \in \mathbb{R}^{d \times d}$, $U^{kse} \in \mathbb{R}^{d \times d}$, $b^{kse} \in \mathbb{R}^d$ are trainable parameters. Note that the two entities share the same knowledge selector parameters.

The results are shown in Table 4. We also list results of **KSM-KS** in the table for comparison. From the results, we can find that:

(1) All the three models with knowledge selectors outperform **KSM-KS**, which illustrates that knowledge selectors could effectively distill not only entity embeddings but also relation embeddings.

(2) **Relation** performs better than **Entity**. There are two potential reasons. On the one hand, relation embeddings are directly used for classification, which would be more suitable for different contexts if they are distilled. However, entity embeddings are used to encode contexts, rather than directly used for classification. Entity knowledge selection has little effect on PPI extraction performance. On the other hand, entity pairs actually exist in instances, while the relations of entity pairs in KBs may not been described in the sequences. So compared with entity knowledge selection, relation knowledge selection is more necessary.

(3) Interestingly, the knowledge selector of relation embeddings (**Relation**) seems more powerful than that of both entity and relation embeddings (**Entity&relation**). This is perhaps that crucial knowledge would be filtered out when knowledge selection is applied on both entity embeddings and relation embeddings.

**Table 4.** Comparison of where the knowledge selector is put on, including nowhere, relation embeddings, entity embeddings and both.

| Position of knowledge selector | P (%) | R (%) | F (%) |
|---|---|---|---|
| **Nowhere (KSM-KS)** | 35.70 | 38.78 | 37.18 |
| **Entity** | 36.81 | 37.74 | 37.27 |
| **Relation (KSM)** | 38.78 | 37.40 | **38.08** |
| **Entity&relation** | 34.83 | 40.28 | 37.35 |

4.6 Effects of Architecture

This section explores the effects of **KSM** architecture. Based on **KSM**, we change some components and get five variants.

**Average**: For the encoded sequences from Transformer encoders, we apply average-pooling rather than a mutual attention to generate context features. Average-pooling takes the mean of each position of $V^{m_1}$ and $V^{m_2}$ as the context features:

$$s^{m_1} = \frac{1}{L}\sum_{i=1}^{L} V_i^{m_1}$$
$$s^{m_2} = \frac{1}{L}\sum_{i=1}^{L} V_i^{m_2} \quad (18)$$

**Max**: For the encoded sequences from Transformer encoders, we apply max-pooling rather than a mutual attention to generate context features. Max-pooling takes the maximum value of each position of $V^{m_1}$ and $V^{m_2}$ as the context features:

$$s^{m_1} = [\max_j V_{1,j}^{m_1}, ..., \max_j V_{i,j}^{m_1}, ..., \max_j V_{d,j}^{m_1}]$$
$$s^{m_2} = [\max_j V_{1,j}^{m_2}, ..., \max_j V_{i,j}^{m_2}, ..., \max_j V_{d,j}^{m_2}] \quad (19)$$

**Separate attention**: For the encoded sequences from Transformer encoders, we apply a separate attention rather than a mutual attention to generate context features from $V^{m_1}$ and $V^{m_2}$. Take $V^{m_1}$ for example:

$$p^{m_1} = \text{soft max}(w^T \tanh(W(V^{m_1})^T + b))$$
$$s^{m_1} = \sum_{i=1}^{L} p_i^{m_1} V_i^{m_1} \quad (20)$$

where $W \in \mathbb{R}^{d \times d}$, $b \in \mathbb{R}^d$ and $w \in \mathbb{R}^d$ are trainable parameters. $s^{m_2}$ is calculated in the same way. Note that the attention parameters are shared between $V^{m_1}$ and $V^{m_2}$.

**One block**: **KSM** has two encoders, each of which has two blocks. This method contains only one block in each encoder.

**Sharing encoder**: **KSM** uses two different encoders corresponding to two entities. This method shares the same encoder between two entities.

**Table 5.** Effects of architecture of **KSM**, including pooling methods (average-pooling, max-pooling), attention methods (separate attention), number of block (one block) and whether two encoders share (sharing encoder) or not.

| Architecture | P (%) | R (%) | F (%) |
|---|---|---|---|
| **Average** | 38.63 | 37.17 | 37.89 |
| **Max** | 35.53 | 37.28 | 36.38 |
| **Separate attention** | 35.05 | 39.24 | 37.02 |
| **One block** | 34.67 | 38.67 | 36.56 |
| **Sharing encoder** | 33.78 | 40.39 | 36.79 |
| **KSM** | 38.78 | 37.40 | **38.08** |

From Table 5, we see that **KSM** achieves the highest $F$1-score, 0.19% higher than **Average**, 1.69% higher than **Max**, 1.06% higher than **Separate attention**, which verifies that the mutual attention pooling is more powerful than other pooling methods in capturing the important context features. Furthermore, the mutual attention could establish the connection between the two entities, making the context features of two entities depend on each other.

Besides, the poor precision of **One block** verifies that one-block sequence encoder is apparently not enough to encode the semantic information in sequences. **Sharing encoder** shares the sequence encoder between two entities, which is a way of reducing the number of parameters but has a negative effect on the precision.

## 5 Discussion

### 5.1 Comparison with Related Work

We fully compare our work with related work by using both Exact Match evaluation measures in Table 6 and HomoloGene evaluation measures in Table 7. In order to make a fair comparison with every system and eliminate the influence of the accumulated errors introduced by different named entity recognition tools, all the systems are reported on the test dataset with the entity annotations recognized by GNormPlus [36] toolkits. We only compare Machine Learning-based (ML) methods without the post-processing rules, and divide these relevant systems into two groups: Machine Learning-based methods with or without knowledge bases, namely ML with KB and ML without KB.

In Table 6, among ML without KB methods, Rios et al. [11] achieve the best $F$1-score 36.77% in Exact Match evaluation measures. Rios et al. [11] take advantage of unlabeled data through learning domain-invariant features with neural adversarial learning.

In Table 7, among ML without KB methods, Tran and Kavuluru [7] obtain the best performance in HomoloGene evaluation measures. They propose a deep convolutional neural network for PPI extraction, which is simple but effective.

Chen et al. [4] use support vector machine with graph kernel to extract PPI. Wang et al. [8] employ recurrent neural network to learn document representations for PPI extraction.

As for ML with KB methods, Zhou et al. [16] leverage prior knowledge about protein-protein pairs with memory networks. They achieve a poor recall since memory networks could not establish direct connections between two arbitrary tokens in a sequence. Our KSM mainly benefits from long-range dependency information and selective prior knowledge. Therefore, KSM keeps the balance between the precision and the recall, and achieves a state-of-the-art $F$1-score.

**Table 6.** Comparison with related work (Exact Match evaluation), including two groups: Machine Learning-based methods with or without KBs.

| Methods | Related work | $P$ (%) | $R$ (%) | $F$ (%) |
|---|---|---|---|---|
| ML without KB | Chen et al.[4] | 34.49 | 32.87 | 33.66 |
| | Tran and Kavuluru [7] | 37.07 | 35.64 | 36.33 |
| | Wang et al.[8] | 9.81 | 50.59 | 16.43 |
| | Rios et al. [11] | 43.98 | 31.59 | 36.77 |
| ML with KB | Zhou et al. [16] | 40.32 | 32.37 | 35.91 |
| | KSM | 38.78 | 37.40 | **38.08** |

**Table 7.** Comparison with related work (HomoloGene evaluation), including two groups: Machine Learning-based methods with or without KBs.

| Methods | Related work | $P$ (%) | $R$ (%) | $F$ (%) |
|---|---|---|---|---|
| ML without KB | Chen et al.[4] | 37.61 | 35.27 | 36.40 |
| | Tran and Kavuluru [7] | 40.07 | 38.63 | 39.33 |
| | Wang et al.[8] | 11.35 | 53.87 | 18.75 |
| ML with KB | Zhou et al. [16] | 42.47 | 34.22 | 37.90 |
| | KSM | 40.81 | 39.54 | **40.16** |

### 5.2 Generalizability Analysis

To verify the generalizability of our method, we conduct an external experiment on BioCreative V Track 3 Chemical Disease Relation (CDR) dataset [40].

A fair comparison is hard to make on the classical PPI tasks (e.g. AIMed [41] and BioInfer [42]), since normalized identifiers of entities are not provided, which are necessary for knowledge extraction from KBs. We process the CDR task data in the same way as Zhou et al. [43].

The results of some baselines mentioned above are compared in Table 8.

As can be seen from the results, **KSM** performs better than **KSM-KS** and **Average**, which shows that both the knowledge selector and mutual attention are effective for CDR extraction. Our proposed method can be generalized well to other relation extraction tasks.

To further demonstrate the effectiveness of our KSM method, the results of the other two state-of-the-art systems are also shown in the Table 8.

**Table 8.** Comparison with some baselines and related work on the CDR dataset to show effects of the knowledge selector and mutual attention.

| Methods | P (%) | R (%) | F (%) |
|---|---|---|---|
| **KSM-KS** | 63.88 | 74.48 | 68.77 |
| **Average** | 65.02 | 75.14 | 69.71 |
| **KSM** | 67.61 | 73.64 | **70.50** |
| **Zhou et al.** [43] | 60.51 | 80.48 | 69.08 |
| **Peng et al.** [44] | 68.15 | 66.04 | 67.08 |

Zhou et al. [43] employ relation embeddings to capture context features, and then concatenate context features with relation embeddings for CDR extraction. However, entity embeddings and knowledge selection are not used. Peng et al. [44] directly extract relations of entity pairs from KBs and represent them as one-hot features for CDR extraction. The one-hot representations could not calculate the semantic relevance between contexts and knowledge. As can be seen from the results, our **KSM** acquires a better $F$1-score than both of them.

5.3 Error Analysis

For 869 PPIs in the test set, 325 are extracted correctly (TP), 544 are not extracted (FN) and 513 are extracted incorrectly (FP). We perform an error analysis on the results of KSM to detect the origins of false positives (FPs) and false negatives (FNs) errors, which are depicted in Fig. 2.

All 513 FPs, with a proportion of 48.53%, are from incorrect classification. Consider the example sentence, "*Both the wild-type ErbB3 and mutant **ErbB3-K/M** proteins transduced signals to phosphatidylinositol 3-kinase, **Shc** and mitogen-activated protein kinases.* (*PMID: 9693119*)", where the two protein entities are shown in bold type. Although there are key words "*mutant*" and "*transduced*" in the sentence, the two entities do not participate in a PPI relation. KSM is confused by these two words and fails to deal with this complex semantic information.

Consider another sentence, "*Mutation of **Gab1** Tyr627 to phenylalanine (YF-Gab1) to prevent the binding of **SHP2** completely prevented the shear stress-induced phosphorylation of **eNOS**, leaving the Akt response intact.* (*PMID: 16284184*)", where the protein pair **Gab1** and **SHP2** is annotated as PPI relation in the test set while the pair **Gab1** and **eNOS** is not. Our model wrongly extracts the latter one as PPI relation because that the contexts of the two pairs are quite similar. The negative instance of pair **Gab1** and **eNOS** could be influenced easily by the positive instance of pair **Gab1** and **SHP2**.

544 FNs come from three factors, false negative entity caused by the GNormPlus toolkits (False negative entity), pre-processing rules (Pre-processing error) and incorrect classification (Incorrect classification). GNormPlus fails to recognize part of entities in the test dataset, which directly leads to 386 FNs with a proportion of 70.96%. We filter out protein pairs distributed across more than two sentences by pre-processing rules in our system, which results in 68 FNs with a proportion of 12.50%. Not surprisingly, KSM incorrectly classifies 90 interacting protein pairs as negative, which only account for 16.54%. In the example sentence, "*Recently, we reported that **ZBP-89** cooperates with histone acetyltransferase coactivator p300 in the regulation of p21(waf1), a cyclin-dependent kinase inhibitor whose associated gene is a target gene of **p53**. (PMID: 11416144)*", the relation of the two entities **ZBP-89** and **p53** is described in a very complicated semantic environment. Detecting PPI relation in such a sentence is difficult for KSM.

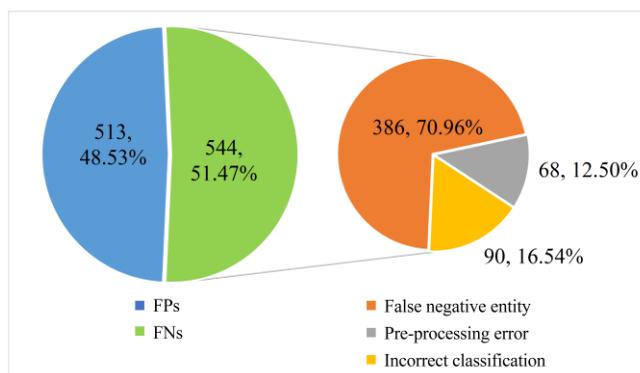

**Fig.2**. The error distribution of all the false positives (FPs) and false negatives (FNs). All the FPs are from incorrect classification, while the FNs can be divided into three groups: False negative entity, Pre-processing error and Incorrect classification.

# 6 Conclusion

This paper develops a PPI extraction model KSM, which could selectively introduce prior knowledge and explicitly capture context information. Two modified Transformers are adapted to effectively compute dependencies between any two context words and encode entity knowledge into contexts. In addition, the mutual attention mechanism is used to establish the connection between the two entities and extract important context features with regard to the protein pair. Furthermore, the knowledge selector is employed to distill relation knowledge based on context features. Experimental results show that KSM could effective take advantage of prior knowledge and contexts for PPI extraction. KSM outperforms the state-of-the-art systems on the BioCreative VI PPI dataset, and can be generalized well to other relation extraction tasks.

## Conflict of Interest

All authors declare no conflict of interest.

## Acknowledgments

This work is supported by Natural Science Foundation of China (No.61772109) and the Ministry of education of Humanities and Social Science project (No. 17YJA740076).